\documentclass[runningheads]{llncs}

\usepackage{graphicx}
\usepackage{verbatim}
\usepackage{makecell}

\usepackage[normalem]{ulem} 
\usepackage[table,xcdraw]{xcolor} 
\usepackage{multirow}
\usepackage{adjustbox}
\usepackage[table,xcdraw]{xcolor}
\usepackage{svg}
\usepackage{hyperref}

\usepackage{booktabs}

\usepackage[T1]{fontenc}
\usepackage{graphicx}
\begin{document}
\title{Assessing the Capability of Large Language Models for Domain-Specific Ontology Generation}

\author{Anna Sofia Lippolis $^{1,2,*}$ 
\and
Mohammad Javad Saeedizade $^{3,*}$ 
\and \\
Robin Keskisärkkä \inst{3}
\and
Aldo Gangemi \inst{1,2}
\and \\
Eva Blomqvist \inst{3}  
\and 
Andrea Giovanni Nuzzolese  \inst{2}  
}
\authorrunning{A. S. Lippolis et al.}

\institute{
  University of Bologna, Italy \\
  \email{annasofia.lippolis2@unibo.it, aldo.gangemi@unibo.it}
  \and
  ISTC-CNR, Italy \\
  \email{andrea.nuzzolese@cnr.it}
  \and
  Linköping University, Sweden \\
  \email{javad.saeedizade@liu.se, robin.keskisarkka@liu.se, eva.blomqvist@liu.se}
}
\maketitle              

\renewcommand{\thefootnote}{*}
\footnotetext{Equal contribution.}
\renewcommand{\thefootnote}{*}
\renewcommand{\thefootnote}{\arabic{footnote}}
\setcounter{footnote}{0}

\begin{abstract}
Large Language Models (LLMs) have shown significant potential for ontology engineering. However, it is still unclear to what extent they are applicable to the task of domain-specific ontology generation. In this study, we explore the application of LLMs for automated ontology generation and evaluate their performance across different domains. Specifically, we investigate the generalizability of two state-of-the-art LLMs—DeepSeek and o1-preview, both equipped with reasoning capabilities—by generating ontologies from a set of competency questions (CQs) and related user stories. Our experimental setup comprises six distinct domains carried out in existing ontology engineering projects and a total of 95 curated CQs designed to test the models’ reasoning for ontology engineering. Our findings show that with both LLMs, the performance of the experiments is remarkably consistent across all domains, indicating that these methods are capable of generalizing ontology generation tasks irrespective of the domain. These results highlight the potential of LLM-based approaches in achieving scalable and domain-agnostic ontology construction and lay the groundwork for further research into enhancing automated reasoning and knowledge representation techniques. 
\end{abstract}
\section{Introduction}
The rapid evolution of large language models (LLMs) has transformed numerous areas of natural language processing as well as knowledge engineering, and ontology engineering is no exception. Ontologies—formal representations of knowledge that enable semantic interoperability—have long been essential in domains ranging from healthcare to environmental science. However, traditional ontology engineering, as shown by requirement-based methods such as eXtreme Design~\cite{presutti2009extreme}, remains a labour-intensive, expertise-driven task, often hindering timely and scalable knowledge modelling. Recent studies, such as the ones by Saeedizade and Blomqvist~\cite{saeedizade2024navigating} and Lippolis et al.~\cite{lippolis2024ontogenia,lippolis2025ontologygenerationusinglarge} suggest a potential for using LLMs in ontology engineering, where LLMs offer a promising addition by automating key aspects of ontology construction through their language understanding and reasoning capabilities.

Building on these earlier studies that showed the feasibility of LLM-supported ontology generation, this paper extends the investigation by suggesting an ontology generation prompt for reasoning models and examining the generalizability of automated ontology generation across diverse and more specific domains. Neither of these aspects were so far addressed in previous work. In particular, we explore the performance of two state-of-the-art LLMs—DeepSeek and OpenAI o1-preview—which are equipped with reasoning capabilities. By leveraging a dataset comprising 95 carefully curated competency questions paired with corresponding user stories, from specific knowledge domains, our approach systematically tests whether these models can generate good ontology drafts based on requirements that capture domain-specific needs.

The contributions of our study are the following: First, we present an automated pipeline for ontology generation, including a prompt strategy, that harnesses the advanced reasoning capabilities of the most recent commercial and open-source LLMs to interpret natural language requirements and suggest corresponding ontological modules. Second, we evaluate this approach across six distinct knowledge domains, providing empirical evidence of its consistent performance and domain-agnostic potential\footnote{Data, code and other material for this study is available at this link: \url{https://github.com/dersuchendee/Domain-OntoGen}}.

The remainder of this paper is organized as follows. Section 2 surveys the relevant literature in LLM-based knowledge engineering. Section 3 establishes the preliminary definitions and concepts used in the remainder of the paper. In Section 4, we describe our methodology, including dataset creation and details about the prompt. Section 5 shows our experimental setup and evaluation framework. Section 6 presents and discusses our results, Section 7 addresses the discussion, and Section 8 focuses on limitations and potential risks. Finally, Section 9 concludes the paper and outlines directions for future research.

\section{Related Work}
Ontology generation based on requirements using large language models has been recently studied. In Benson et al.\cite{benson2024my}, the authors present a study on producing outputs that are consistent with BFO through the chat interface of ChatGPT (with GPT-4). However, this work was based on a few examples and was solely focused on LLMs with non-reasoning capabilities, noting reasoning models needed a new, ad-hoc analysis. In Saeedizade and Blomqvist~\cite{saeedizade2024navigating}, LLM-generated ontologies from a set of competency questions from a semantic web course were compared to student performance in that course. The domain of those ontology stories was limited to ontology stories designed for that specific course. The study by Lippolis et al.~\cite{lippolis2024ontogenia} on ontology generation was based on the African Wildlife Ontology, a gold standard outlined in Potoniec et al.~\cite{potoniec2020dataset}.  Additionally, Doumanas et al.~\cite{doumanas2025fine} fine-tuned LLMs to generate ontologies by considering the domain of the ontology. However, the comparison of its performance between different domains is not mentioned in the study. In the same way, the work by Lippolis et al.~\cite{lippolis2025ontologygenerationusinglarge} addresses LLM-based ontology generation on a dataset with 100 CQs of different domains, but doesn't address the LLM's performance on the involved domains and focus on three ontology homework from a semantic web course, the one presented in~\cite{saeedizade2024navigating}.
For what concerns studies about domain-specific applications with LLMs, Fathallah~\cite{fathallah2024llms4life} focuses on ontology generation in life sciences, Huang et al.~\cite{huang2024large} in drinking water distribution network and Xu et al.~\cite{xu2024knowledge} evaluate retrieval augmented generation for E-commerce and~\cite{alharbi2024exploring} on drug indication, assessing the capability of LLMs in generating knowledge graphs from text using predefined ontologies. Val-Calvo et al. propose a pipeline to automate ontology development and knowledge graph generation with GPT-4 from CSV files from the e-commerce domain to streamline the engineering process in enterprise settings~\cite{VALCALVO2025104042}. From these studies, it is possible to conclude there is no systematic evaluation of one method across different ontology domains of varying complexity, especially with different LLMs with reasoning capabilities.

\section{Preliminaries}
In this section, we present and define terminology that will be used throughout this paper, building on the one already outlined in Lippolis et al.~\cite{lippolis2025ontologygenerationusinglarge}. 

\textbf{Ontology Generation}: We define ontology generation as the process of creating formal representations of knowledge by identifying the necessary classes, properties, and axioms to capture domain-specific concepts and their relationships. This process can be performed manually or supported by automated techniques, such as LLMs, to draft ontologies from natural language inputs in the form of requirements.

\textbf{Modelling Competency Question}:
Following the definitions outlined in previous work~\cite{lippolis2025ontologygenerationusinglarge}, a CQ is considered modelled in an ontology if and only if a SPARQL query exists to extract the answer of the CQ from the ontology irrespective of the quality of the ontology modelling or adherence to optimal modelling practices.

\textbf{Minor issue}: Similar to ~\cite{lippolis2025ontologygenerationusinglarge}, for a pair ontology and a competency question, if the ontology includes all necessary elements (named classes or properties) except for only one object property or only one data property, and adding this single element would make the CQ modelled, this is considered a minor issue in modelling the CQ.

\textbf{Reasoning Capability of LLMs}:
The reasoning capability of LLMs refers to their ability to generate logically structured and coherent responses by recognizing and using patterns learned from large amounts of text. This definition includes simulating various forms of reasoning—such as deducing, generalizing from examples, and drawing analogies—often by processing information step-by-step (a process sometimes called chain-of-thought, which is natively employed in models like o1-preview). Importantly, LLM reasoning is based on statistical patterns rather than true human understanding or deliberate logic.

\section{Methodology}
In this section, we outline the methodology employed in this study. We begin by describing the dataset creation process and its statistical properties, followed by the explanation of the prompting technique used in this work. Finally, we present the pipeline for ontology generation.

\subsection{Dataset Creation}

To evaluate ontologies generated by LLMs with reasoning capabilities, we constructed a dataset comprising $95$ competency questions (CQs) with $12$ ontology stories across $6$ distinct domains. Each dataset entry includes: (i) a CQ, (ii) a user story (providing the contextual background), (iii) the binary label of the difficulty level of the CQ (easy or hard), and (iv) the domain of the ontology. Table \ref{tab:domain_stats} shows statistics for the dataset.

\subsubsection{Dataset composition}
 The CQs were extracted from four sources: \emph{Onto-DESIDE} \footnote{\url{https://ontodeside.eu}} (domain: Circular Economy), the \emph{Polifonia Project} (domains: Music and Events)\cite{de2023polifonia}, the \emph{WHOW Project} (domain: Water and Health)~\cite{lippolis2025water} and \emph{AquaDiva} (domains: Microbe Habitat and Carbon and Nitrogen Cycling)\cite{algergawy2024towards}. In particular, these derive from carefully curated large-scale projects, with the fourth one being a semi-automatically annotated dataset.

\subsubsection{CQ classification}
To ensure balanced and comparable evaluations, each CQ was classified into one of two categories, ``Easy'' and ``Hard'', based on the complexity required for its formal representation. An LLM, o1-preview, was used to assign these labels by estimating the difficulty level of classes and properties needed to model the CQ. Specifically, if a CQ required at most $2$ classes and $1$ property (either a data or an object property), it was classified as ``Easy''; otherwise, it was classified as ``Hard''. The labels were then checked manually by an ontology engineer. This labelling strategy was implemented to maintain a consistent distribution of CQ difficulties across all domains.

\begin{table}[htbp]
\centering
\caption{Domain Statistics for Competency Questions.}
\resizebox{0.7\textwidth}{!}{

\begin{tabular}{l @{\extracolsep{20pt}} c c c}
\toprule
\textbf{Domain} & \textbf{Total CQs} & \textbf{Easy} & \textbf{Hard} \\
\midrule
Circular Economy              & 16 & 5  & 11 \\
Music (Bells, Organs, Tunes)    & 16 & 10 & 6  \\
Events                        & 18 & 8  & 10 \\
Microbe Habitat               & 15 & 8  & 7  \\
Carbon and Nitrogen Cycling   & 15 & 4  & 11 \\
Water and Health              & 15 & 7  & 8  \\
\midrule
\textbf{Total}                & 95 & 42 & 53 \\
\bottomrule
\end{tabular}
}
\label{tab:domain_stats}
\end{table}

\subsubsection{Manual annotations}
The labelling of the dataset for the verification of CQs has been carried out manually by two ontology engineers. Following the error classification outlined in previous work~\cite{lippolis2025ontologygenerationusinglarge}, the output ontologies were considered to be modelled, not modelled, or not modelled due to a minor error.

\subsection{Prompting Techniques}
A prompt serves as an instruction for LLMs to generate responses to a given input. The prompt employed in this study follows a few-shot prompting technique, with some examples, building on and distinguishing it from previous approaches which mainly employed a decomposed prompting technique~\cite{saeedizade2024navigating,lippolis2024ontogenia,lippolis2025ontologygenerationusinglarge}. This choice is motivated by the observation that reasoning models have been shown in Lippolis et al.~\cite{lippolis2025ontologygenerationusinglarge} to perform more effectively with minimal instructions rather than being explicitly guided to think step by step. After trials with and without a few-shot example, the performance of the few-shot has proven more accurate than the other.
The prompt is structured into three key components:  

\begin{itemize}
    \item \textbf{Instruction}: A description of the task (ontology generation) and its objective (ontology must model the input CQ).  
    \item \textbf{Example}: A Turtle syntax sample ontology model corresponding to a CQ and its associated ontology story, illustrating an ideal response.  
    \item \textbf{Input}: A designated section where a new CQ and ontology story are inserted, prompting the LLM to generate an ontology model accordingly.
\end{itemize}


\subsection{Ontology Generation}
We used the Independent Ontology Generation method described in~\cite{lippolis2025ontologygenerationusinglarge}.
In this approach, each CQ and its associated ontology story are provided to an LLM through a prompt to generate the corresponding ontology. This method treats each CQ as an independent unit, enabling a precise evaluation of the ontology generation process on a per-question level. By handling CQs in isolation, the impact of the settings can be assessed.

\section{Experiment Setup}

In this section, we describe the experimental setup. To run the experiments, we used Azure API to call o1-preview and the original DeepSeek API with default hyperparameters. The prompt with a fixed example (few-shot) is appended with ontology story and CQ and sent to the API to get the resulting output file. The results are saved in a Turtle file.

\textbf{LLMs}:
In the present study, we have selected two large language models with reasoning capabilities: DeepSeek and OpenAI o1-preview. This selection was motivated by the desire to perform a comparative analysis between the current state-of-the-art closed-source and open-source LLMs. The OpenAI o1-preview model was accessed through the Microsoft Azure API, whereas DeepSeek was used according to the specifications detailed in its official documentation\footnote{\url{https://api-docs.deepseek.com/}}.

\textbf{Prompt}:
The prompt used in this work is available in the supplementary materials on GitHub\footnote{\url{https://github.com/dersuchendee/Domain-OntoGen}}. Initially, several prompt configurations were evaluated, including versions with and without examples (i.e., few-shot and zero-shot settings). The results demonstrated that prompts incorporating examples consistently outperformed those without, justifying their use. Furthermore, the impact of varying the complexity of the examples was examined in a separate experiment involving multiple ontology generation tasks. In the interest of conciseness, the detailed results of these additional experiments are not reported here.

\textbf{Hyperparameters}:
The default hyperparameters for the LLMs, including temperature and penalty, were employed throughout this study.

\section{Results}
Overall, both the o1-preview and DeepSeek models exhibited high and comparable accuracy. Initially, nine and ten CQs remained unmodeled for the o1-preview and DeepSeek models, respectively, out of a total of 95 CQs. However, when minor modelling issues are excluded, the number of unmodeled CQs decreases to eight for the o1-preview model and five for the DeepSeek model. 

Table~\ref{tab:mistakes} provides a breakdown of mistakes on both easy and hard CQs by domain. Regarding the o1-preview scores, performance was consistent across all domains, suggesting that the ontology generation method is broadly applicable rather than being limited to a specific domain. With slight differences, although DeepSeek generally demonstrated similar consistency, the scores for the Events domain were noticeably lower, indicating a need for further investigation into the model's performance in that particular domain.

\begin{table}[ht]
\centering
\caption{Count of errors in modelling for each LLM, divided into Easy and Hard competency questions grouped by ontology domain}
\label{tab:mistakes}
\setlength{\tabcolsep}{12pt} 
\resizebox{0.95\textwidth}{!}{
\begin{tabular}{|l|cc|cc|c|cc|cc|c|}
\hline
\multirow{2}{*}{Category} & \multicolumn{5}{c|}{o1-preview} & \multicolumn{5}{c|}{DeepSeek} \\
\cline{2-11}
 & \multicolumn{2}{c|}{Not modelled} & \multicolumn{2}{c|}{Contains a minor issue} & Total & \multicolumn{2}{c|}{Not modelled} & \multicolumn{2}{c|}{Contains a minor issue} & Total \\
\cline{2-3}\cline{4-5}\cline{7-8}\cline{9-10}
 & Easy & Hard & Easy & Hard &  & Easy & Hard & Easy & Hard &  \\
\hline
Circular Economy            & 0 & 0 & 0 & 2 & 2 & 0 & 0 & 0 & 0 & 0 \\
\hline
Music                       & 0 & 1 & 1 & 0 & 2 & 2 & 0 & 0 & 0 & 2 \\
\hline
Events                      & 0 & 0 & 1 & 1 & 2 & 1 & 2 & 0 & 2 & 5 \\
\hline
Microbe Habitat             & 0 & 0 & 0 & 0 & 0 & 0 & 0 & 1 & 0 & 1 \\
\hline
\multicolumn{1}{|l|}{\makecell{Carbon and\\Nitrogen Cycling}} & 0 & 0 & 0 & 1 & 1 & 0 & 0 & 0 & 1 & 1 \\
\hline
Water and Health            & 0 & 0 & 0 & 2 & 2 & 0 & 0 & 1 & 0 & 1 \\
\hline
\end{tabular}
}
\end{table}

\section{Discussion}

Our results reveal a performance trade-off between the two models. Specifically, while o1-preview tends to commit more ``minor'' errors, DeepSeek is prone to a higher incidence of incomplete modellings. When we consider cases with ``minor'' mistakes equivalent to complete modellings, both models demonstrate similar performance across the domains. This finding suggests that differences in error types may not translate into significant disparities in overall outcomes.

\paragraph{Domain generalizability.} Our work extends previous findings by Lippolis et al.~\cite{lippolis2025ontologygenerationusinglarge} by showing that reasoning models working for a specific domain can generalize effectively across other datasets. Therefore, the improvements achieved in earlier studies are not limited to a single domain, affirming the broader applicability of the developed techniques.

\paragraph{Reasoning models architecture.} One possible explanation for the similar outputs observed from both DeepSeek and o1-preview is their reasoning capabilities. It is plausible that these models benefit from mutual fine-tuning or shared architectural strengths. The collective evidence points to a significant enhancement in reasoning with respect to models with non-reasoning capabilities, which appears to underlie their comparable performance.

\paragraph{Difficulty levels of the requirements.} Furthermore, the analysis indicates that both models handle hard and easy CQs with similar performance. This observation challenges the common assumption that increased question complexity necessarily results in lower scores. Instead, our findings suggest that the models are robust enough to process questions of varying complexity without a marked performance penalty. Nonetheless, this is limited to only one evaluation criterion.

\paragraph{Error analysis.} Finally, a closer look at the errors reveals that the majority occur solely within the Events domain. The corresponding CQs in this category are much shorter than those in other domains, but there are multiple user stories that serve as requirements. This pattern could suggest that longer, more descriptive CQs could provide additional context that aids the models in processing and reasoning, leading to improved performance, especially if multiple scenarios are provided. Therefore, the lack of standardization in requirements is also fundamental when dealing with automated techniques.

Overall, these insights highlight the importance of both model architecture and the design and selection of requirements in achieving balanced and generalizable performance.

\section{Limitations and Risks}
One potential risk associated with this study is dataset leakage, which may impact the quality and reliability of the LLM-generated ontology. Specifically, in certain domains, an LLM may have encountered parts of the dataset during pretraining, while in others, it relies solely on its reasoning capabilities. To the best of our knowledge, the CQs and ontology modules were not publicly available in a single dataset, reducing the likelihood that LLMs were trained on these elements in direct succession as consecutive tokens. Another limitation of this work is the exclusive use of reasoning-based LLMs. These models, while beneficial for ontology generation tasks, are generally resource-intensive and less widely available compared to standard LLMs. Additionally, the scope of this study is restricted to only two LLMs, which may limit the generalizability of the findings.

\section{Conclusion}

In this study, we investigated the potential of two LLMs, DeepSeek and o1-preview, with reasoning capabilities for automated ontology generation across six diverse domains. Our experiments, conducted on a dataset of 95 competency questions paired with user stories, reveal that both LLMs similarly exhibit robust and consistent performance across all domains and difficulty levels of the CQs.
Overall, this work lays a foundation for further advancements in automated ontology engineering, highlighting the potential of LLMs to deliver scalable, domain-agnostic knowledge representation. Future research will build on these results to explore additional LLMs.

\begin{credits}
\subsubsection{\ackname}
This project has received funding from the European Union’s Horizon Europe research and innovation programme under grant agreements no. 10105- 8682 (Onto-DESIDE) and 101070588 (HACID), and is supported by the strategic research area Security Link. Additional financial support to this project was provided by NextGenerationEU under NRRP Grant agreement n. MUR IR0000008 - FOSSR ( CUP B83C220- 03950001 ).
This work was also supported by the PhD scholarship ``Discovery, Formalisation and Re-use of Knowledge Patterns and Graphs for the Science of Science'', funded by CNR-ISTC through the WHOW project (EU CEF programme - grant agreement no. INEA/CEF/ICT/ A2019/2063229). Finally we thank OpenAI's Researcher Access Program Grant for the API credits.

\subsubsection{\discintname}
The authors have no competing interests to declare that are
relevant to the content of this article.
\end{credits}
%
%
%
\bibliographystyle{splncs04}
\bibliography{mybibliography}

\end{document}